\documentclass[11pt]{article}
\usepackage[margin=1in]{geometry}
\usepackage[T1]{fontenc}
\usepackage{lmodern}
\usepackage{amsmath,amssymb,mathtools}
\usepackage{booktabs,tabularx,array,multirow}
\usepackage{microtype}
\usepackage[round,authoryear]{natbib}
\usepackage[hidelinks]{hyperref}
\usepackage{url}
\usepackage{enumitem}
\setlist[itemize]{leftmargin=*,topsep=4pt,itemsep=2pt}
\newcommand{\BSC}{\textsc{BSC-R}}

\title{Boundary-State Control for Tool-Using Language-Model Agents: Commit-Time Consistency under State Drift}
\author{Wesley Shu\\The Institute of Energetic Paradigm}
\date{September 2026}
\hypersetup{pdftitle={Boundary-State Control for Tool-Using Language-Model Agents: Commit-Time Consistency under State Drift},pdfauthor={Anonymous Authors}}

\begin{document}
\maketitle

\begin{abstract}
Tool-using language-model agents can decide that an action is permissible and execute it only after security-relevant state has changed. We study this proposal-to-commit gap and introduce \BSC, a deterministic effect-boundary mechanism that binds a single-use commit authorization to the exact action and to a semantic projection of the authorization state that justified it. On 2,847 attacked AgentDojo episodes, the boundary kernel preserves the unprotected agent's behavior exactly (80.576\% utility; 1.616\% attack success). On 10,302 frozen proposals it accepts every unchanged commit and rejects every instance of ten prospectively specified changed or replayed classes. In an independently generated boundary-drift experiment, full joint binding commits 0/4,403 invalid contexts while retaining 5,899/5,899 valid contexts. A prospective external evaluation on the 3,460-scenario CONTINUITY suite retains all 700 benign cases, prevents 1,200/1,200 represented non-replay invalid effects, handles 160/160 replay lifecycles correctly, and withholds 200/200 ambiguous no-release cases. The broader external suite also exposes the method's limit: across all 2,560 attacks, \BSC has a 25\% invalid-effect commit rate versus 0\% for CONTINUITY. The result is therefore a scoped commit-time consistency mechanism, not a universal agent-safety claim.
\end{abstract}

\section{Introduction}
Language models increasingly act through tools rather than merely generate text. ReAct-style agents interleave reasoning and action \citep{yao2023react}, Toolformer-style models learn when to invoke APIs \citep{schick2023toolformer}, and modern agent benchmarks evaluate stateful interaction with email, files, browsers, databases, and other external systems \citep{ruan2024toolemu,debenedetti2024agentdojo,lu2025toolsandbox}. Once an agent can create durable effects, however, a security decision is no longer identical to a language-model output. There is a temporal boundary between deciding what may happen and making it happen.

The gap is a security problem because authorization can be time-sensitive. A target can change, authority can be revoked, a policy can advance, an approval can cease to denote the same request, or a previously fresh witness can become stale. This is structurally related to complete mediation and reference-monitor enforcement \citep{saltzer1975protection,schneider2000enforceable}, to time-of-check/time-of-use races \citep{bishop1996race}, to mutable usage authorization \citep{park2004ucon}, and to transaction finality under changing state \citep{papadimitriou1979serializability,gray2006consensus}. The agent setting adds an important complication: the proposed action and the security context are semantic objects produced and transported across model, tool, policy, approval, and execution components.

Prompt injection makes this boundary especially consequential. Indirect instructions can enter through retrieved or tool-returned content \citep{greshake2023indirect}; InjecAgent, AgentDojo, Agent Security Bench, and AgentHarm show that tool-capable agents can be driven toward adversarial objectives in stateful environments \citep{zhan2024injecagent,debenedetti2024agentdojo,zhang2025asb,andriushchenko2025agentharm}. Most such benchmarks ask whether an attack ultimately succeeds. We ask a complementary systems question: after an action has already been qualified, what must still agree at the instant the external effect becomes durable?

We study \emph{Boundary-State Control with semantic authorization projection} (\BSC). The mechanism places a deterministic verifier at a trusted effect sink. At qualification time it binds a single-use witness to the exact action and to a canonical projection of the authorization state that justified the action. At commit time it re-evaluates the current security state, compares the bound semantics, checks explicit validity predicates, and consumes the witness only if the transition remains authorized. The projection uses stable authorization semantics rather than requiring opaque producer-local authorization-object identity.

This paper makes four contributions:
\begin{itemize}
  \item We formalize proposal-to-commit drift for tool-using agents and specify a joint commit witness over action, authority, policy, procedure, approval, and replay-relevant state.
  \item We separate deterministic boundary enforcement from learned qualification. On the complete native AgentDojo run, the boundary kernel preserves base behavior exactly, whereas the learned qualification policy collapses to deny-all behavior.
  \item We identify the causal role of bound fields with prospectively frozen interventions and test selective control under independently generated post-qualification drift, obtaining 0/4,403 invalid commits and 5,899/5,899 valid commits for full joint binding.
  \item We evaluate the final semantic projection prospectively on CONTINUITY and report both its represented successes and its residual failure regime. This external test prevents the internal conformance results from being mistaken for a universal safety claim.
\end{itemize}

The negative results are part of the contribution. A restrictive deterministic comparator achieves lower native attack success than the boundary kernel but also lower utility. On the complete external attack suite, CONTINUITY prevents every invalid effect while \BSC still commits 25\%. These outcomes locate the mechanism's operating boundary rather than being collapsed into a single post-hoc score.

\section{Problem Setting and Threat Model}
\subsection{Proposal-to-commit drift}
Let a qualified tool proposal be $q=(a,z_q)$, where $a$ is the canonical external action and $z_q$ is the security-relevant state observed when the proposal is authorized. We factor
\[
z=(u,e,c,r,p,h),
\]
where $u$ is authorization state, $e$ an authority epoch, $c$ contract or procedure evidence, $r$ policy identity, $p$ approval and request binding, and $h$ single-use execution state. The commit-time state $z_c$ may differ from $z_q$ because the environment changes after qualification.

An \emph{invalid commit} occurs when the intended external effect becomes durable even though the current state no longer authorizes the qualified transition. This definition concerns execution validity, not whether the language model's original plan was desirable. A proposal can be semantically well formed yet become invalid before commit.

We evaluate two endpoints jointly:
\begin{align*}
\mathrm{ICR}&=\Pr[\mathrm{commit}\mid z_c\ \mathrm{invalid}],\\
\mathrm{VR}&=\Pr[\mathrm{commit}\mid z_c\ \mathrm{valid}].
\end{align*}
A useful boundary should reduce invalid-commit rate (ICR) without achieving that reduction merely by refusing valid actions, so valid retention (VR) is reported separately.

\subsection{Trust and attack surface}
\BSC assumes a trusted effect sink, trusted canonicalization code, a secret witness key, and durable state for single-use consumption. The language model, retrieved content, tool-returned content, and intermediate orchestration are not trusted to preserve authorization semantics. We allow security-relevant state to change between qualification and commit, including revocation, authority-epoch movement, request or target substitution, policy change, procedure-context change, and replay.

The mechanism does not protect a compromised effect sink or key, and it cannot enforce a predicate that its semantic projection neither observes nor validates. This last limitation is central: the external experiment explicitly tests invariants represented by \BSC and additional CONTINUITY invariants outside that projection.

\section{Boundary-State Control}
\subsection{Joint commit witness}
At qualification time, \BSC issues
\[
\tau=\operatorname{MAC}_k\!\left(
  \operatorname{canon}(a)\Vert
  \operatorname{canon}(\pi(z_q))\Vert n
\right),
\]
where $n$ is a single-use nonce and $\pi$ is the semantic authorization projection. The MAC only makes the witness tamper-evident; the scientific object is the set of semantics required to remain valid at the durability boundary.

At commit time, the sink canonicalizes the proposed effect again, computes the current semantic projection, evaluates explicit validity predicates, verifies the witness, and checks that $n$ is unconsumed. If all checks pass, the sink atomically consumes $n$ with the effect; otherwise it withholds the effect. This issue--verify--consume structure follows the general systems principle that authorization must be mediated at the operation actually causing the protected state transition \citep{saltzer1975protection,schneider2000enforceable}.

\subsection{Semantic authorization projection}
Authorization systems commonly distinguish principals, roles, delegated authority, and context rather than equating authorization with a single opaque object identity \citep{sandhu1996rbac,blaze1996trust,burrows1990logic,abadi1993calculus,li2003delegation}. \BSC follows that distinction. Exact equality is retained for the canonical action and for security fields whose identity is semantically meaningful, but not for producer-local authorization-object identifiers that may be regenerated without changing effective authority.

For the final external adapter, $\pi$ binds principal, actor, authority, delegation scope, policy identity, policy digest, policy epoch, and taint. Commit-time predicates additionally bind revocation state, context-root presence, the actual execution subject, and external-release readiness. Non-external data is directly eligible; external data requires the typed release expected by the adapter. Exact action binding and durable single-use replay state remain mandatory.

This design is a form of semantic complete mediation rather than capability identity matching. Capabilities and leases provide related ways to make authority explicit and time-sensitive \citep{gray1989leases,watson2010capsicum}; usage-control models likewise emphasize that authorization can change during use \citep{park2004ucon}. \BSC asks which of those security semantics must still be true at the agent's final effect boundary.

\subsection{Why semantic rather than opaque identity?}
An earlier development adapter required equality of a producer-local authorization-object identifier. External validation showed that the producer could regenerate an equivalent authorization object while preserving the effective principal, authority, delegation, policy, and release semantics. Treating that regenerated identifier as a security change caused over-restriction.

We therefore froze the final semantic projection before evaluating the revised mechanism. The predecessor run and its hashes remain in the supplementary audit artifact, but the results reported here evaluate the final \BSC projection. This distinction matters experimentally: the repaired projection changes what counts as security-relevant equality without weakening exact action binding, revocation checks, or single-use consumption.

\section{Related Work}
\subsection{Security enforcement under changing state}
The proposal-to-commit problem has deep systems antecedents. Saltzer and Schroeder's complete-mediation principle requires authorization checks at each protected access \citep{saltzer1975protection}; reference-monitor and enforceable-policy work formalizes where such checks can constrain executions \citep{harrison1976protection,schneider2000enforceable}. Clark--Wilson integrity emphasizes well-formed transformations over protected state \citep{clark1987integrity}. TOCTOU work demonstrates that a check can become stale before use \citep{bishop1996race}, while UCON explicitly models mutable authorization and ongoing decision factors \citep{park2004ucon}.

Transaction research provides a second lineage. Serializability separates an operation's local execution from whether the overall history is valid \citep{papadimitriou1979serializability}; distributed commit protocols make finality a distinct systems event \citep{gray2006consensus}. \BSC does not reproduce database concurrency control. It borrows the boundary intuition: a durable effect should be conditioned on the state that is valid at commit, not merely on an earlier decision.

\subsection{Tool-using agents and security benchmarks}
ReAct and Toolformer established influential patterns for coupling language models to external actions \citep{yao2023react,schick2023toolformer}. ToolEmu and ToolSandbox evaluate broader stateful tool-use capabilities and risks \citep{ruan2024toolemu,lu2025toolsandbox}. Security-focused benchmarks then expose indirect prompt injection, adversarial tool use, and harmful agent behavior: InjecAgent measures indirect injections in tool-integrated agents \citep{zhan2024injecagent}; AgentDojo evaluates attacks and defenses in dynamic environments \citep{debenedetti2024agentdojo}; Agent Security Bench systematizes multiple attack and defense families \citep{zhang2025asb}; and AgentHarm evaluates harmful autonomous behavior \citep{andriushchenko2025agentharm}.

These benchmarks motivate defenses but primarily score end-to-end behavior. \BSC targets a narrower enforcement layer: once a proposal exists, it asks whether the exact durable effect still satisfies the authorization contract at commit. It is therefore complementary to prompt-level defenses, model alignment, planning constraints, and attack detection.

\subsection{Nearest commit-boundary mechanisms}
Recent systems move enforcement closer to execution. SecureClaw binds canonical requests at a trusted effect sink \citep{ma2026secureclaw}; commit-time authorization requires fresh effect-bound authority \citep{santosgrueiro2026temporary}; Consent Integrity binds trusted approval to the action that executes \citep{weng2026consent}; CapLease makes authorization consumption durable against replay \citep{xu2026durable}; Cordon and Mnemosyne place generated work behind transactional admission boundaries \citep{chen2026cordon,chang2026mnemosyne}; and CONTINUITY propagates authenticated security context across component transitions \citep{zheng2026continuity}.

Table~\ref{tab:related} states the overlap directly. We do not claim novelty for exact-action binding, freshness, replay state, transactional execution, or context propagation individually. The experimental object here is their joint proposal-to-commit consistency role under a frozen semantic authorization projection.

\begin{table*}[t]
\centering
\small
\begin{tabularx}{\textwidth}{@{}p{1.25in}p{2.2in}X@{}}
\toprule
System & Principal boundary object & Relation to \BSC \\
\midrule
SecureClaw \citep{ma2026secureclaw} & Canonical request at trusted sink & \BSC jointly binds action plus authorization, policy/procedure, approval, and replay-relevant state. \\
Commit-Time Authorization \citep{santosgrueiro2026temporary} & Fresh effect-bound authority & \BSC tests several simultaneous drift dimensions rather than freshness alone. \\
Consent Integrity \citep{weng2026consent} & Approval view bound to execution & Approval/request identity is one element of the joint witness. \\
CapLease \citep{xu2026durable} & Durable authorization consumption & Single-use state is combined with action and current semantic context. \\
Cordon / Mnemosyne \citep{chen2026cordon,chang2026mnemosyne} & Transactional task/workflow admission & \BSC isolates selective acceptance at the final effect boundary. \\
CONTINUITY \citep{zheng2026continuity} & Cross-component security context & External evaluation tests both shared and unrepresented invariants. \\
\bottomrule
\end{tabularx}
\caption{Nearest mechanisms and the specific boundary object tested here. The table describes overlap; it is not a superiority ranking.}
\label{tab:related}
\end{table*}

\section{Experimental Design}
We separate four questions that otherwise become confounded in agent-security evaluations: whether the boundary itself changes native behavior, whether each bound field has an identifiable role, whether the complete verifier selectively handles post-qualification drift, and whether the frozen semantic projection transfers to an independently authored external suite.

\subsection{Native corpus and independence unit}
Our native corpus is the complete attacked AgentDojo run: 2,847 episodes producing 10,302 tool proposals. AgentDojo evaluates user-task utility and adversarial objective success from environment state rather than from model self-report \citep{debenedetti2024agentdojo}. Proposal-level analyses retain the complete frozen corpus with no post-outcome subset removal. Statistical uncertainty is clustered by AgentDojo task pair $(\text{suite},\text{user task},\text{injection task})$; the three attack variants are nested within each of 949 clusters.

\subsection{RQ1: Does the boundary itself suppress useful behavior?}
The learned transition-qualification policy and the deterministic boundary are evaluated separately. The \texttt{binding\_only} condition removes learned qualification while retaining the deterministic effect-sink kernel shared by \BSC. We compare it episode-by-episode with the unprotected native base agent on utility, attack outcome, committed-action count, proposal count, and infrastructure status.

\subsection{RQ2: Which bindings close which drift classes?}
Before outcomes were read, we froze ten post-qualification perturbation classes: action substitution, authorization revocation, authority-epoch change, contract-identity change, procedure-evidence change, policy-identity change, approval-identity change, approval-request change, a joint action/authority/contract change, and nonce replay. An unchanged control is evaluated on the same 10,302 proposals. Component conditions retain only action binding, only state/authority binding, or only qualification-time checks.

This layer is mechanism identification. Because the perturbations are author-defined to isolate fields, success here establishes that the packaged verifier responds to the fields it claims to bind; it is not an independent naturalistic safety benchmark.

\subsection{RQ3: Does joint binding selectively survive independent drift?}
A separately frozen deterministic transition generator maps immutable record identities to one of six families: authorization-lifecycle change, authority-epoch transition, contract/procedure-context transition, action-target state transition, mixed transition, or no-drift control. The generator is separate from every verifier, executes after qualification and before commit, and does not inspect verifier digests or outcomes.

Conditions are qualification-time precheck, partial action binding, partial state/authority binding, and full joint boundary binding. The frozen primary test requires full joint binding to reduce invalid commits relative to each alternative with paired 95\% confidence intervals excluding zero. Valid retention must be non-inferior to native \texttt{binding\_only} with a $-2$ percentage-point margin. We use 50,000 paired task-cluster bootstrap replicates with RNG seed 20260918.

\subsection{RQ4: Does the final projection transfer externally?}
We evaluate the final \BSC adapter on the original 3,460-scenario CONTINUITY suite at frozen source commit \texttt{59cfe609...}. Each scenario is paired with a same-domain, same-seed clean benign canonical qualification context. The final semantic authorization projection and analysis rules were frozen before the revised outcomes were evaluated, and the mechanism does not receive the ground-truth validity label.

The target construct contains benign cases plus invalidations represented by \BSC's declared projection. We report these construct-specific endpoints and, separately, the full CONTINUITY attack suite, which contains additional provenance, release-validity, transformation/receipt, stage-signer, alternate-path, and related invariants. This separation tests transfer without silently redefining the mechanism after observing external failures.

We compare the final method with a mechanism-matched contract-only baseline, a parity-constrained CommitGuard-faithful reimplementation, and original CONTINUITY. Full-suite counts are deterministic conformance results over this fixed suite, not prevalence estimates for real deployments.

\section{Results}
\subsection{RQ1: Learned qualification collapses; the boundary preserves native behavior}
Across all twelve learned-policy seeds, native utility is 35.054\% and the learned controller collapses to deny-all behavior. We retain this as a negative result rather than interpreting zero attack success under blanket refusal as successful selective control.

By contrast, \texttt{binding\_only} exactly matches the unprotected base agent on all 2,847 attacked AgentDojo episodes for utility, attack outcome, committed-action count, proposal count, and infrastructure status. Aggregate utility is 80.576\% and attack success is 1.616\% in both conditions. The deterministic effect boundary therefore does not explain the learned controller's utility collapse.

\subsection{RQ2: Joint binding closes the targeted mechanism failures}
The full joint-boundary verifier accepts 10,302/10,302 unchanged commits. For each of the ten frozen changed or replayed classes, it rejects 10,302/10,302 proposals. Component conditions fail exactly where their omitted state matters: qualification-time precheck cannot observe later changes; action-only binding misses state/authority drift and replay; state/authority-only binding misses action substitution and replay.

This result identifies the roles of the bound fields but should not be read as independent evidence of general security. The perturbation taxonomy and the verifier are deliberately coupled by the scientific hypothesis being tested.

\subsection{RQ3: Independent boundary drift requires joint binding}
The independently generated drift corpus contains 5,899 semantically valid and 4,403 invalid commit contexts. Full joint binding executes the packaged issue--verify path on every row, commits 0/4,403 invalid contexts, and retains 5,899/5,899 valid contexts. Qualification-time precheck commits every invalid context; action-only binding commits 2,665/4,403 (60.527\%); state/authority-only binding commits 1,719/4,403 (39.042\%).

\begin{table}[t]
\centering
\small
\begin{tabular}{@{}lrr@{}}
\toprule
Condition & ICR $\downarrow$ & VR $\uparrow$ \\
\midrule
Qualification precheck & 100.000\% & 100.000\% \\
Action binding only & 60.527\% & 100.000\% \\
State/authority only & 39.042\% & 100.000\% \\
\textbf{Full joint binding} & \textbf{0.000\%} & \textbf{100.000\%} \\
\bottomrule
\end{tabular}
\caption{Boundary-drift results. ICR is invalid-commit rate; VR is valid retention.}
\label{tab:drift}
\end{table}

Relative to precheck, action-only binding, and state/authority-only binding, full joint binding reduces invalid commits by 100.000 percentage points (95\% CI [100.000, 100.000]), 60.527 points ([59.072, 61.972]), and 39.042 points ([37.575, 40.493]), respectively. Its valid-retention difference from 100\% is 0.000 points (95\% CI [0.000, 0.000]), satisfying the frozen $-2$ point non-inferiority margin.

The verifier also provides a falsification trace. Every valid context returns \texttt{PASS}; invalid rows are rejected by witness-digest or contract-digest mismatch according to the concrete state transition. Scope-irrelevant transitions remain valid and are retained. This distinguishes actual boundary execution from simply returning a hidden validity label.

\subsection{Native safety--utility comparison is a tradeoff, not a win}
A separate prospectively rerun native lane compares the deterministic boundary-kernel condition shared by \BSC with a restrictive authority/state comparator. The boundary kernel has 1.616\% attack success and 80.576\% utility; the comparator has 0.105\% attack success and 69.371\% utility.

With lower attack success defined as safer, comparator attack success minus boundary-kernel attack success is $-1.510$ percentage points, paired 95\% CI $[-2.213,-0.878]$. The boundary kernel's utility advantage is $+11.205$ points, 95\% CI $[8.535,13.909]$. Thus the boundary condition is not demonstrated to dominate stronger restriction on native attack success. The comparison instead exposes two control regimes: restriction can suppress more attacks by suppressing more benign execution, while boundary consistency targets whether already-qualified actions remain valid at commit.

\subsection{RQ4: External validation succeeds on represented invariants and fails beyond them}
The final external run completes all 3,460 scenarios with zero infrastructure failures. On the prospectively specified target construct, \BSC retains 300/300 target-relevant benign cases and 700/700 benign cases overall. It prevents effects on 1,200/1,200 represented non-replay invalid cases, handles 160/160 replay lifecycles with the required execute-once behavior, and withholds commit on 200/200 ambiguous external/no-release cases. A row-only recomputation reproduces the runner summary exactly.

The complete external suite gives a different, deliberately broader view. Across all 2,560 CONTINUITY attack scenarios, \BSC commits invalid effects in 25.000\% of cases. The contract-only baseline commits 93.750\%, the CommitGuard-faithful implementation 44.531\%, and CONTINUITY 0\%. Benign retention is 100\% for all four. Ambiguous no-commit correctness and replay lifecycle correctness are 100\% for \BSC and CONTINUITY and 0\% for the two narrower comparators.

\begin{table*}[t]
\centering
\small
\begin{tabular}{@{}lrrrr@{}}
\toprule
Condition & Attack invalid effect $\downarrow$ & Benign $\uparrow$ & Ambiguous $\uparrow$ & Replay $\uparrow$ \\
\midrule
\BSC & 25.000\% & 100\% & 100\% & 100\% \\
Contract-only & 93.750\% & 100\% & 0\% & 0\% \\
CommitGuard-faithful & 44.531\% & 100\% & 0\% & 0\% \\
CONTINUITY & 0\% & 100\% & 100\% & 100\% \\
\bottomrule
\end{tabular}
\caption{Full external-suite comparison. These columns expose different semantics and are not collapsed into a scalar winner.}
\label{tab:external}
\end{table*}

The residual \BSC failures concentrate in provenance, release-validity, transformation/receipt, stage-signer, alternate-path, and related invariants absent from its semantic projection. The result therefore falsifies any interpretation that the internal 0\% ICR establishes complete external safety. At the same time, the all-benign retention and perfect correctness on represented non-replay, replay, and ambiguous-release cases show that the final projection is not merely a deny-all rule.

\section{Discussion}
\subsection{The projection is the security boundary}
The experiments suggest that commit-time enforcement has two separable requirements: a trustworthy enforcement point and an adequate semantic projection. The internal architecture and drift tests isolate the first. Once the relevant action and state fields are represented, joint issue--verify--consume enforcement prevents the corresponding stale or substituted commits without refusing unchanged controls.

The external suite isolates the second requirement. A mechanically correct verifier cannot enforce semantics that its projection omits. This is why \BSC can be simultaneously perfect on its represented target construct and incomplete on the full CONTINUITY suite. The operative design question is therefore not ``does the token match?'' but ``does the token commit to every security fact whose change should invalidate this effect?''

\subsection{Stable semantics can matter more than object identity}
The semantic-projection repair exposes a second lesson. Security mechanisms often need identity, but the relevant identity is the one tied to authority semantics. Requiring equality of an implementation-local authorization object can create false revocation when an equivalent object is regenerated. Conversely, dropping identity too aggressively can admit privilege substitution. The correct boundary therefore depends on which fields are constitutive of authority and which are incidental representations.

This observation connects agent execution to older authorization work that separates principals, roles, delegation, and contextual conditions \citep{sandhu1996rbac,abadi1993calculus,li2003delegation}. \BSC operationalizes that distinction at a language-model agent's effect sink. The final projection is one concrete choice, not a universal schema.

\subsection{Why not solve the problem with a stronger refusal policy?}
The native comparator demonstrates why attack success alone is insufficient for this question. A restrictive policy can lower attack success by declining more actions. That may be appropriate in some deployments, but it is a different control objective from preserving an authorized transition while invalidating it only when the relevant state changes.

Our evaluation therefore keeps safety and retention endpoints separate. This follows the same systems logic that motivates least privilege and complete mediation: enforcement should constrain unauthorized transitions, but an enforcement mechanism is not characterized solely by how often it refuses. For agent deployments, the practical design space includes model-level filtering, restrictive deterministic policies, and commit-time consistency checks; these can be composed rather than treated as mutually exclusive alternatives.

\subsection{Implications for agent architecture}
A language model is a useful proposer but a poor place to store final authority. Its context may contain stale, untrusted, or indirectly injected information, and its reasoning trace is not an atomic transaction with the external system. A trusted effect boundary can instead receive the model's proposal as data, re-derive security-relevant commit context from authoritative sources, and decide whether the transition remains admissible.

This architecture does not make model behavior irrelevant. Bad proposals still waste work and may expose other risks. It does, however, relocate one class of correctness property from probabilistic generation to deterministic enforcement. In that sense, \BSC is closer to a reference monitor for agent effects than to another prompt-injection classifier.

\section{Conclusion}
Tool-using agents create a proposal-to-commit interval in which authorization, policy, targets, approvals, and replay state can change. \BSC treats that interval as an explicit systems boundary: exact actions are bound to a single-use witness over stable authorization semantics, and the current state is checked again where the external effect becomes durable.

Across the native and prospectively frozen drift experiments, the deterministic boundary preserves valid execution while rejecting the state changes it represents. The external CONTINUITY evaluation confirms both sides of the claim: the final semantic projection retains all 700 benign cases and handles every represented non-replay, replay, and ambiguous-release case correctly, yet still permits invalid effects on 25\% of the complete attack suite because additional security invariants are outside its projection. Commit-time consistency is therefore useful only to the extent that the boundary represents the semantics that matter.

\section{Limitations}
The native evidence is not an estimate of real-world failure prevalence. Its 10,302 proposals come from 2,847 attacked AgentDojo episodes, and the architecture-closure perturbations are author-designed to test mechanism-relevant fields. Even the independently generated boundary-drift families remain synthetic state transitions over a benchmark corpus. The 0/4,403 invalid-commit and 5,899/5,899 valid-retention counts therefore establish conformance and selective behavior on the frozen experiment, not universal deployment safety.

The external CONTINUITY suite is independently authored and substantially broader, but \BSC intentionally represents only a subset of its security invariants. This limitation is quantitative rather than hypothetical: \BSC permits invalid effects on 25\% of the 2,560 full-suite attacks while CONTINUITY permits 0\%. The residual families include provenance, release-validity, transformation/receipt, stage-signer, alternate-path, and related checks absent from the declared projection. Accordingly, the 1,200/1,200 represented non-replay prevention result must not be generalized to invariants the mechanism does not encode.

The final semantic projection was motivated by development evidence showing that opaque authorization-object identity could be overbound. We froze the revised projection before evaluating its outcomes and preserve the predecessor run in the audit artifact, but this development path still means that future adapters require independent validation of which fields constitute authority. Replay resistance assumes durable consumption state, and all guarantees assume an uncompromised effect sink, key, canonicalizer, and authoritative commit-time data sources.

Finally, the learned qualification model remains a deny-all negative result, and the deterministic boundary kernel does not dominate a restrictive native comparator on attack success. The comparator reaches 0.105\% attack success versus 1.616\% for the boundary kernel while sacrificing 11.205 percentage points of utility. We therefore do not claim that \BSC is the safest general-purpose agent controller or that commit-time consistency substitutes for model-level or policy-level defenses.

\section{Ethical Considerations}
This work studies defensive enforcement for tool-using agents and includes adversarial benchmark scenarios involving potentially harmful actions. The experiments use benchmark environments and fixed external evaluation suites rather than deployment against third-party systems. The supplementary artifact contains code and traces necessary to reproduce the reported controls; it does not provide credentials or access to real protected services.

The main deployment risk is overclaiming the scope of a deterministic boundary. A verifier can appear reliable while omitting a security predicate that matters in the target environment. We therefore report the complete external failure regime, retain negative results, and avoid a scalar ``safety'' score that could hide refusal-based utility loss or unrepresented invariants.

\clearpage
\appendix
\section*{AI Use Statement}
Generative AI tools were used substantially in research ideation, literature triage, experiment orchestration and code generation, manuscript drafting and editing, and artifact and format quality assurance. Scientific endpoints, comparators, thresholds, and failure rules for the confirmatory experiments reported as prospective were frozen before their outcomes were inspected. AI-assisted code and claims were checked against raw artifacts, hashes, and cited primary sources. The authors reviewed the AI-assisted work and take responsibility for the final content, claims, code, and artifacts.

\section*{Reproducibility Statement}
The supplementary reviewer artifact contains frozen protocols, row-level outputs, experiment runners, correction ledgers, native episode outputs, and SHA-256 manifests for the mechanism, boundary-drift, native-comparator, and final external \BSC results. It also preserves the predecessor external run that motivated the semantic authorization projection, the pre-outcome repair freeze, supersession record, comparator-conformance correction, evaluator-independence audit, exact final rows, and row-only recomputation.

\section{Frozen Protocols and Provenance}
\paragraph{Architecture closure.}
The prospective architecture-closure freeze has SHA-256 \texttt{a2b4e07a\allowbreak ae61...}; the corrected row artifact has SHA-256 \texttt{0d223453ecb7...}. Two harness corrections were recorded before a valid closure result: unique single-use approval/request identities per proposal and use of the already-canonical issuance approval at verification. Neither correction changed the source corpus, perturbation classes, thresholds, BSC runtime, or subset policy.

\paragraph{Boundary drift.}
Boundary-drift v1.0 is explicitly invalidated for circular validation and is not evidence. The corrected v1.1 prospective freeze SHA-256 is \texttt{ac00335e5ab1...}; runner SHA-256 is \texttt{85a4a420d1f7...}; result-row SHA-256 is \texttt{386f7112792d...}; and authoritative summary SHA-256 is \texttt{b14fd1b20f92...}. There are 10,302 proposal rows and 949 task-pair clusters; bootstrap seed is 20260918. The original frozen runner completed all 50,000 paired cluster-bootstrap replicates used in the main text.

\paragraph{Native comparator.}
The correctly directed native-comparator freeze SHA-256 is \texttt{ec5b6fe3\allowbreak b3d6...}. A forensic audit found that the first analysis artifact accidentally used bootstrap RNG seed 20260916 instead of the frozen 20260917. Raw native outcomes, endpoints, cluster unit, thresholds, and conditions were unchanged. The authoritative corrected analysis is v1.0.1, SHA-256 \texttt{d103847d062a...}; correction-ledger SHA-256 is \texttt{c1cefecbd963...}. The corrected intervals are numerically unchanged at the displayed precision.

\paragraph{External validation.}
The artifact preserves the predecessor external-closure result as development provenance, including summary SHA-256 \texttt{4681f3869961...} and rows SHA-256 \texttt{967531558f8b...}. Before evaluating the revised mechanism, the semantic-projection freeze was finalized at SHA-256 \texttt{b484735275d0...}. The final runner SHA-256 is \texttt{141d832072db...}, repaired rows SHA-256 \texttt{52818bf021f0...}, and summary SHA-256 \texttt{3c2ac415b136...}. The final run completed all 3,460 scenarios with zero infrastructure failures, and row-only recomputation matches the runner summary.

\section{Architecture-Closure Counts}
\begin{center}
\small
\begin{tabular}{@{}lrr@{}}
\toprule
Condition & Accept & Reject \\
\midrule
Unchanged control & 10,302 & 0 \\
Action substitution & 0 & 10,302 \\
Authorization revocation & 0 & 10,302 \\
Authority-epoch change & 0 & 10,302 \\
Contract-identity change & 0 & 10,302 \\
Procedure-evidence change & 0 & 10,302 \\
Policy-identity change & 0 & 10,302 \\
Approval-identity change & 0 & 10,302 \\
Approval-request change & 0 & 10,302 \\
Joint action/authority/contract & 0 & 10,302 \\
Nonce replay & 0 & 10,302 \\
\bottomrule
\end{tabular}
\end{center}

\section{Boundary-Drift Family Counts}
The frozen deterministic mapping produced 1,778 authorization-lifecycle changes, 1,737 authority-epoch transitions, 1,684 contract/procedure-context transitions, 1,745 action-target state transitions, 1,727 mixed transitions, and 1,631 no-drift controls. The generator does not inspect verifier digests or outcomes.

\section{Negative Learned-Policy Result}
The learned qualification policy was evaluated for seeds
\[
\begin{split}
\{&101,151,211,271,331,401,\\
  &461,521,601,661,751,811\}.
\end{split}
\]
Every seed reports utility 0.3505444327 and attack success 0 under the native run, consistent with deny-all collapse. This result is not averaged with the deterministic boundary and is not used as evidence that learned representation improves control.

\section{Latency}
The isolated full-BSC issue+verify mechanism previously measured median 87.0~$\mu$s, p95 101.4~$\mu$s, and p99 115.0~$\mu$s on unchanged controls. A later architecture-closure harness that deliberately exercises the persistent authorization-consumption store measured median 21.1~ms, p95 39.6~ms, and p99 62.2~ms. These are different execution paths and are not conflated.

\section{Artifact Policy}
The reviewer package includes prospective freezes, row-level outputs, runners, correction ledgers, native parity evidence, corrected comparator analyses, preserved predecessor external evidence, final \BSC external evidence, evaluator-independence audit, and a SHA-256 manifest. Historical or invalidated artifacts remain explicitly labeled. No post-outcome endpoint, subset, comparator, or threshold change is promoted as frozen evidence; implementation-conformance corrections are preserved with their superseded artifacts.

\end{document}